\documentclass[letterpaper, 10 pt, conference]{ieeeconf}  
\makeatletter
    \let\NAT@parse\undefined
\makeatother

\usepackage[square, numbers, sort&compress]{natbib}

\IEEEoverridecommandlockouts                              

\usepackage{diagbox} 
\usepackage{booktabs} 
\usepackage{graphicx}
\usepackage{amsmath}
\usepackage{booktabs}
\usepackage{wrapfig}
\usepackage[table]{xcolor} 
\usepackage{caption} 
\usepackage{hyperref}

\newcommand{\mypara}[1]{\par\noindent\textbf{{#1}}}

\title{\LARGE Enabling Multi-Robot Collaboration from Single-Human Guidance}

\author{Zhengran Ji$^{1}$, Lingyu Zhang$^{1}$, Paul Sajda$^{2}$, Boyuan Chen$^{1}$
\\
\textcolor{orange}{\href{http://www.generalroboticslab.com/HUMAC}{www.generalroboticslab.com/HUMAC}}
\thanks{This work is supported by ARL STRONG program under awards W911NF2320182 and W911NF2220113. Authors are from $^{1}$Duke University and $^{2}$Columbia University.}
}

\begin{document}

\makeatletter
\let\@oldmaketitle\@maketitle%
\renewcommand{\@maketitle}{\@oldmaketitle%
    \centering

    \includegraphics[width=\textwidth]{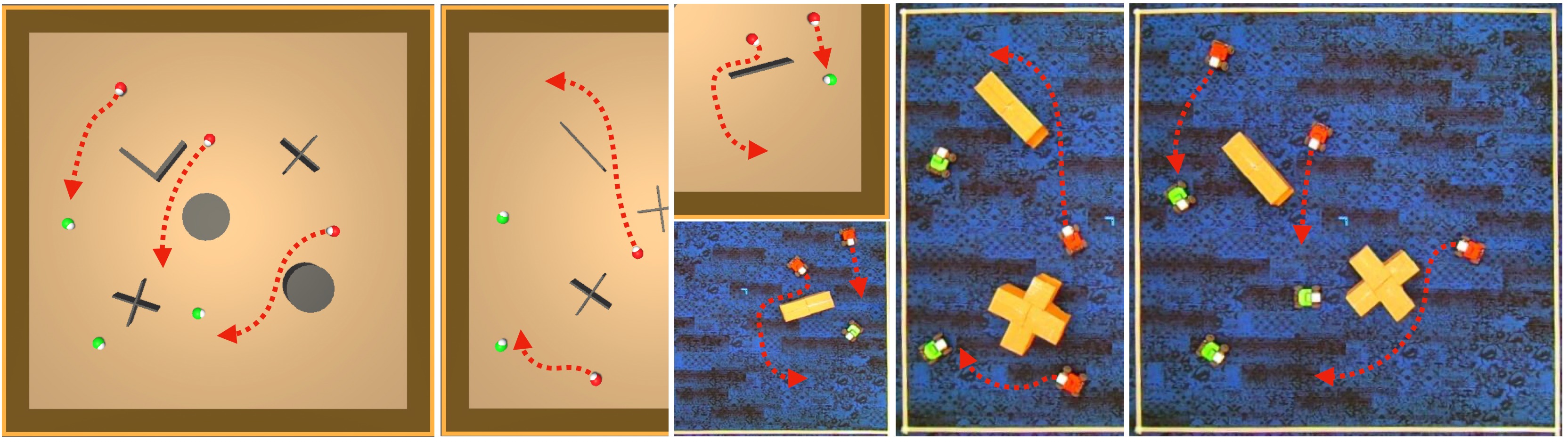}
    \captionof{figure}{Our framework enables multi-robot collaboration in dynamic multi-agent hide-and-seek tasks from single-human guidance. The best policy achieves an average seeker success rate of 84.2\% in simulation and 80\% in real-world experiments in a challenging 3 seekers vs 3 hiders setting with random map layouts. In comparison, the baseline policy has only 36.4\% in simulation and 55\% in real-world. Interesting collaborative behaviors among seekers are observed during deployment, such as strategically navigating to anticipate and intercept hiders or effectively blocking key paths as a team.}
    \label{fig:teaser}
    \vspace*{-5mm}
}
\makeatother

\maketitle
\thispagestyle{empty}
\pagestyle{empty}
\setcounter{figure}{1}

\begin{abstract}

Learning collaborative behaviors is essential for multi-agent systems. Traditionally, multi-agent reinforcement learning solves this implicitly through a joint reward and centralized observations, assuming collaborative behavior will emerge. Other studies propose to learn from demonstrations of a group of collaborative experts. Instead, we propose an efficient and explicit way of learning collaborative behaviors in multi-agent systems by leveraging expertise from only a single human. Our insight is that humans can naturally take on various roles in a team. We show that agents can effectively learn to collaborate by allowing a human operator to dynamically switch between controlling agents for a short period and incorporating a human-like theory-of-mind model of teammates. Our experiments showed that our method improves the success rate of a challenging collaborative hide-and-seek task by up to 58$\%$ with only 40 minutes of single-human guidance. We further demonstrate our findings transfer to the real world by conducting multi-robot experiments.

\end{abstract}

\section{Introduction}

Humans, as inherently social beings, thrive in an interactive world where collaboration enables the accomplishment of remarkable achievements \cite{huamncolab}. Our ability to work together has been instrumental in building civilizations surpassing other species' capabilities, including those physically stronger or larger than us. 
This naturally raises the question: can the collaborative capability of humans be realized in multi-agent autonomous systems?


Learning collaborative AI agents is a significant yet challenging problem with broad potential applications across various domains. In real-world settings, collaborative AI agents can be deployed in areas such as drone swarms, autonomous vehicles, warehouse automation, and manufacturing systems to optimize tasks like search and rescue operations, traffic management, and production processes. However, developing such agents remains challenging, as it requires each agent's ability to infer their teammates' intentions and goals through long-horizon and spatial reasoning and to take appropriate actions to collaborate effectively. Most existing real-world collaborative multi-agent systems are tailored to specific scenarios and cannot generalize effectively when the number of agents or environmental layout changes, limiting their applicability in dynamic, real-world environments.

Recent advancements in multi-agent reinforcement learning (MARL) \cite{maddpg,maac,mappo,qmix,vdn} have demonstrated success in collaborative tasks. Collaborative behaviors often emerge as agents maximize a shared reward \cite{tampuu2017multiagent, socialdilemma, emergethrucompete, capture}. However, there is no guarantee when or if these collaborative behaviors will appear\cite{bakeremergent}. Moreover, MARL typically requires millions to billions of environment interactions due to the inherent low sample efficiency of reinforcement learning and the added complexity of non-stationarity\cite{bakeremergent}. Designing MARL algorithms to learn collaboration explicitly is still challenging, as providing quantitative definitions of collaborations in reward engineering is difficult. On the other hand, multi-agent imitation learning (MAIL) provides a more efficient and controllable framework to learn collaborative behaviors by imitating demonstrations from a group of collaborating experts \cite{le2017coordinated, copulas, driving, multigail, scalable}. Yet, this approach requires a large amount of data from multiple experts, with the number of experts typically scaling linearly with the number of agents in the team, making it both costly and challenging to obtain.


We observe that humans naturally possess the ability to embody different roles within a team \cite{qin2025perception,qin2025physiologically}. Rather than relying on a team of experts, in this paper, we introduce a mechanism (Fig.~\ref{fig:mainfigure}) that enables a team of agents to learn collaborative behaviors from the guidance of a single human. The framework effectively simulates a collaborative team demonstration by allowing one human to dynamically switch between controlling different agents through interventions. Additionally, we develop a computational framework with effective policy representations, inspired by human Theory of Mind \cite{wellman2004scaling, apperly2009humans}, to model teammate behaviors and efficiently learn from human interventions.


We evaluate our method (Fig.~\ref{fig:mainfigure}) on a challenging multi-agent hide-and-seek task involving partial visual observations. With 10 to 40 minutes of single-human guidance, a robot team exhibits strong collaborative behaviors such as ambushing and encircling. Experiment results demonstrate that our method outperforms baselines by up to 58\% with 40 minutes of single-human guidance, and these findings successfully transfer to real-world multi-robot systems. Our method further generalizes robustly under stochastic environment layouts and different numbers of teammates and opponents. Additionally, the high success rate of our computational framework indicates the effectiveness of our policy representations of both the agent itself and its teammates. Furthermore, our analysis and ablation studies provide a deeper understanding of our single-human guidance mechanism and teammate modeling policy representations.


\section{Related Work}

\subsection{Multi-Agent Reinforcement Learning (MARL)} 
Recent advancements in Deep Reinforcement Learning \cite{dqn, ppo, ddpg, sac} have significantly progressed in training multi-agent systems from environment interactions and reward feedback. Many algorithms designed initially for single-agent RL have been adapted to solve multi-agent problems in continuous \cite{mappo, maddpg, maac} and discrete action settings.\cite{vdn, qmix}. However, MARL often assumes collaborative behavior might emerge as agents maximize their total expected return \cite{reviewcoop}, without guaranteeing when or if it will occur \cite{bakeremergent}. MARL inherits the low sample efficiency from single-agent RL but is also more challenging due to additional non-stationarity. Even in environments with simplified state representations, state-of-the-art MARL methods require millions to billions of interactions \cite{bakeremergent}. Furthermore, explicitly imposing collaborative behaviors in MARL is challenging to scale and generalize, as it demands extensive reward engineering and quantitative definitions of collaborations. In contrast, our approach leverages human expertise to guide agents in learning collaboration through imitation.

\subsection{Multi-Agent Imitation Learning (MAIL)} MAIL typically relies on demonstrations of collaborative behavior, assuming the availability of coordinated trajectories from individual expert-controlled agents. Prior work \cite{le2017coordinated} has jointly learned a latent coordination model and individual policies from demonstrations of a group of experts. Another method models the dependence among individual agent policies through a copula function \cite{copulas}. A multi-agent actor-critic algorithm has also been introduced \cite{multigail} using a generalized version of inverse RL. At the same time, a combination of GAIL and parameter-sharing TRPO has been applied to multi-agent autonomous driving \cite{driving}. A Bayesian formulation for MAIL has also been proposed to improve sample efficiency, \cite{yang2020bayesian}.

However, curating such datasets is challenging, requiring multiple experts to coordinate simultaneously. This approach demands significant human effort \cite{driving}, often resulting in inconsistent demonstration quality due to individual differences. Most existing works are tested in environments simple enough to derive analytical collaborative policies \cite{multigail, copulas}, which limits their scalability to more complex real-world problems. Our approach differs by requiring only a single human to control multiple agents in sequence, effectively simulating time-wide collaboration without the need for multiple human experts. Additionally, we evaluate our method in a more realistic setting where the environment is random and dynamic, making it extremely difficult to derive optimal collaborative behaviors analytically.

\subsection{Theory of Mind Inspired Machine Learning}
Theory of Mind (ToM) \cite{wellman2004scaling, apperly2009humans}, the ability to understand and model the mental states and intentions of others plays a critical role in enabling effective teamwork. Prior research has explored how such capabilities can be adapted to learning agents \cite{agentsmodelagents}. One approach \cite{machinetom} uses a ToM network to acquire a prior model of agents' behaviors, enabling the predictions of the agents' characteristics and ``mental states". Another method \cite{cogmachinetom} uses a cognitive model based on instance-based learning theory to enhance generalization across different settings. A vision-based model \cite{chen2021visualtob} has been shown to allow AI observers to predict complex behaviors of another agent purely through visual input and further used for visual hide-and-seek in 1v1 setting \cite{chen2021visual}. Additionally, a symmetric scenario \cite{symmetricmachinetom}, where agents freely interact, has been introduced to demonstrate that RL agents modeling mental states can significantly improve performance. Inspired by ToM, we have explored various design choices within our proposed framework to enable our agents to model teammate behaviors effectively.

\begin{figure*}[t!]
    \centering
    \includegraphics[width=1\textwidth]{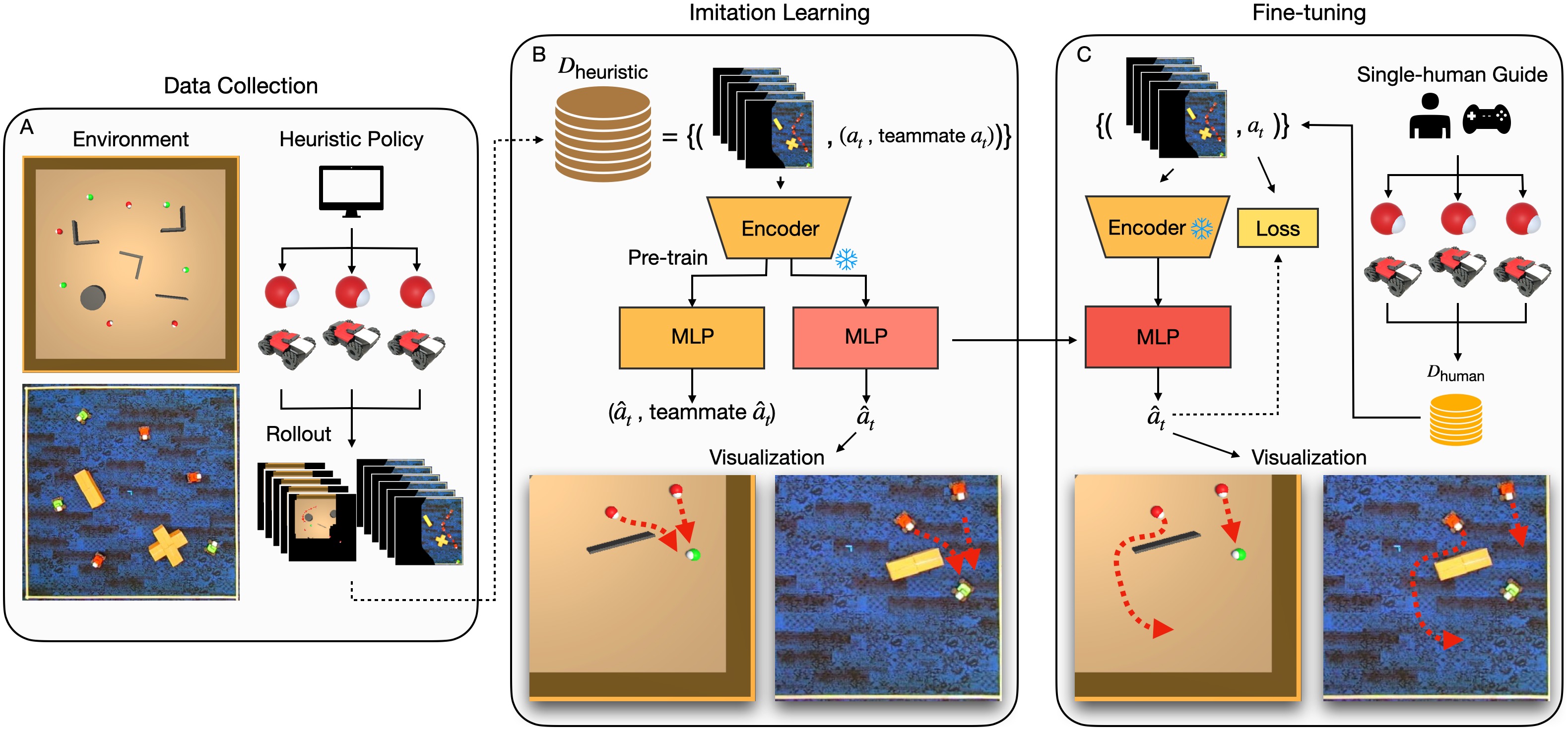}
    \caption{Our framework uses a single human to guide multiple agents in learning collaborative behaviors. (A) Data Collection: We use a predefined heuristic policy to collect a dataset $D_\text{heuristic}$. (B) Imitation Learning (IL): We pre-train the encoder by predicting the whole team's actions with a single agent's observations and only update the MLP during IL. The IL policy exhibits no collaboration. (C) Fine-tuning: We collect single-human intervention data $D_\text{human}$ and fine-tune the policy with a frozen encoder. The fine-tuned policy demonstrates effective collaboration during testing.}
    \label{fig:mainfigure}
    \vspace*{-5mm}
\end{figure*}

\section{Method}

\subsection{Task Settings}

\mypara{Multi-Agent Visual Hide-and-Seek} While it is challenging to quantify collaboration in multi-agent systems directly, task performance can serve as an indirect measure when collaboration is necessary. Multi-agent hide-and-seek \cite{hideandseek, chen2021visual, bakeremergent} (Fig.~\ref{fig:mainfigure}A) is a classic predator-prey task where seekers, shown in \textcolor{red}{red}, cooperate to catch hiders, shown in \textcolor{green}{green}. In this work, we design a version of the task where effective collaboration among seekers is crucial for success. Unlike past works that simplify the setting with low-dimensional state observations\cite{bakeremergent} or ground-truth observations of the opponent team \cite{bakeremergent, chen2021visual}, our agents rely solely on partial visual observations. Our focus is on fostering collaborative behavior among seekers.

Both agent types navigate the environment at constant speeds; however, to make collaboration among seekers essential, hiders are assigned a higher speed than seekers. The task happens in a bounded square-shaped arena with random obstacles. At the start of each episode, agents are spawned at random positions near the arena's boundary to ensure they are far apart. Seekers have a partially observable top-down field of view similar to SLAM used in robotics \cite{smith1986representation}, where the history of the static surroundings is accumulated. The poses of other seekers are also rendered on the map to inform their teammates' poses. The hider agent has a circular, egocentric top-down field of view, slightly smaller than the seeker's. This design is necessary. Otherwise, we have observed that the hiders, who already move faster, would easily escape the seeker's view, making it extremely difficult for the seekers even to spot them. The seekers are successful if they catch all hiders within a predefined time limit.





\begin{figure}[t!]
    \centering
    \includegraphics[width=0.45\textwidth]{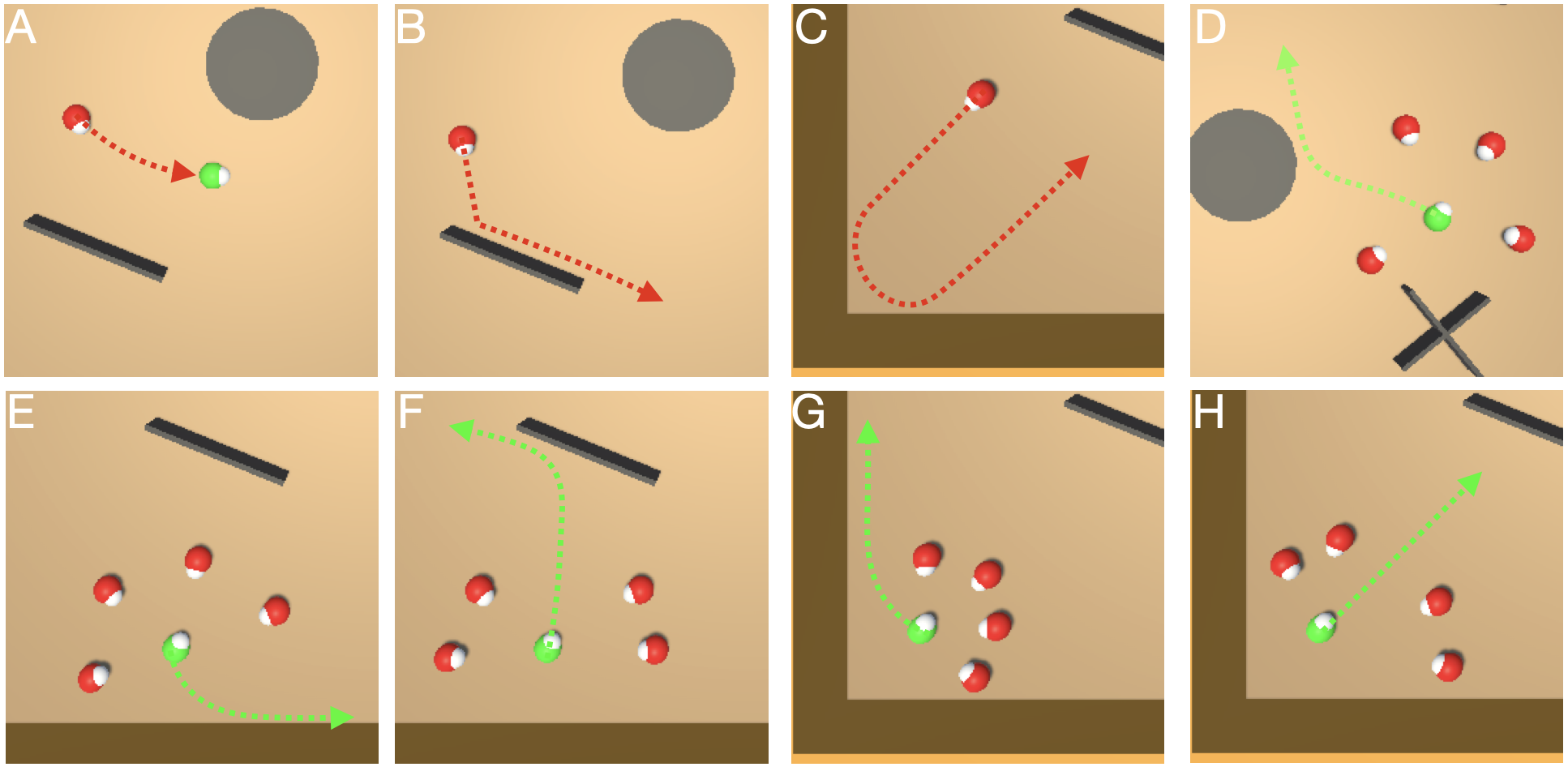}
    \vspace*{-2mm}
    \caption{Heuristic policy. (A-C) Seeker chases hider, avoids obstacles, and avoids walls. (D-H) Hider runs away from multiple seekers.}
    \label{fig:Heuristic}
    \vspace*{-7mm}
\end{figure}

\mypara{Hider Policy} Since we focus on developing policies for the seekers, the hiders should follow a robust policy that maximizes their chances of escape in complex scenarios, such as being cornered by multiple seekers near walls or obstacles. Our designed heuristic hider policy aims to be flexible and responsive with geometric computations, with collaborative hider behaviors left for future exploration. As shown in Fig.~\ref{fig:Heuristic}D-H, when multiple seekers are within the hider's field of view, and the hider is near a wall or corner, hiders can escape by selecting the optimal route to avoid capture. Our setting ensures the task remains challenging, as non-collaborative seekers who keep chasing closest hiders can only achieve a 36\% success rate in a 3v3 setting.

\mypara{Seeker Default Policy} As a baseline for non-collaborative behavior, we implement a heuristic-based policy for the seeker. If one or more hiders are detected within the observable range, the seeker will pursue the nearest hider (Fig.~\ref{fig:Heuristic}A-C). 

\mypara{Seeker Learned Policy} We formulate the seeker's decision-making process as the following learning problem. The policy controls waypoint-based navigation, where at each time step $t$, the policy $\pi$ predicts the next target location and orientation $a_t$ given the current visual observation $s_t$: $\pi(s_t) = a_t$. During execution, we assume that the agents' low-level motion planner can accurately navigate to the desired location.

\subsection{Single-Human Guidance}
To maximize the impact of single-human guidance, we designed the single-human guidance mechanism with the following considerations:

\mypara{Dynamic Switching} An efficient way of grounding human expertise is to collect real-time demonstrations. Our interface allows the human operator to freely switch between seekers to intervene in their current policies at any time, addressing the challenge of teleoperating multiple agents.

\mypara{Minimizing Cognitive Load} Given the limited cognitive capacity of humans, asking a human to control multiple agents simultaneously at all times would quickly lead to cognitive overload. Our key idea is that human interventions are not always required. Most of the time, simple heuristic policies can guide the agents effectively after appropriate human interventions. In some cases, effective collaboration depends on other agents functioning as expected. Therefore, we ask the human operator to intervene only when necessary while the remaining seekers follow their original policies. Additionally, when the human chooses not to provide input, all agents maintain their original policies, ensuring coherent and effective team behavior. Our approach also implicitly mitigates the non-stationarity challenge.

\subsection{Data Collection}

To learn a collaborative seeker policy, we collect two datasets. The first is the heuristic-controlled dataset, $D_\text{heuristic}$, where the heuristic policy controls all seeker agents. The second is the human intervention dataset, $D_\text{human}$, where a single human guides the seekers using the mechanism described above. Notably, we found that a small size of $D_\text{human}$, compared to $D_\text{heuristic}$, was sufficient for achieving good performance. At each time step $t$, we collect the current visual observations $s_t$ and pose $(p_t, o_t)$ of each seeker agent. We then pair $s_t$ with $(p_{t+1}, o_{t+1})$ as the agent's action $a_t$.

\subsection{Grounding Human Guidance to Enable Collaborations}

A straightforward approach to ground human guidance in seeker policies is to directly train an imitation learning policy using the human guidance data $D_\text{human}$. However, the limited amount of human data can lead to overfitting. To overcome this, we first train the seeker on abundant $D_\text{heuristic}$ obtained by rolling our heuristic policies without human involvement. We then fine-tune the policy using the much smaller $D_\text{human}$.

\mypara{Imitation Learning(IL)} We use ResNet-18 \cite{He_2016_CVPR} as the backbone for our policies due to its strong performance in vision tasks. To incorporate trajectory history, we stack the past $N$ frames as input. We empirically find that $N=5$ in simulation and $N=3$ in the real-world yield optimal results. To leverage the power of convolutional networks and balance the weights across input dimensions, we introduce a binary mask as an additional channel to the RGB frame. This mask has a value of one at the location of the corresponding seeker agent, with all other pixels set to zero. Rather than encoding low-dimensional poses separately, we integrate positional information directly into the visual input. Since our goal is not to develop advanced imitation learning algorithm backbones, we adopt a standard imitation learning framework, optimizing a mean squared error loss $\mathcal{L}$ with $D_\text{heuristic}$:
\begin{equation}
    \mathcal{L}(\pi) = \frac{1}{|D_{\text{heuristic}}|}\sum_{(s_t, a_t)\in D_{\text{heuristic}}} \left\| \pi(s_t) - a_t \right\|^2
\end{equation}

\begin{figure}[t!]
    \centering
    \includegraphics[width=0.48\textwidth]{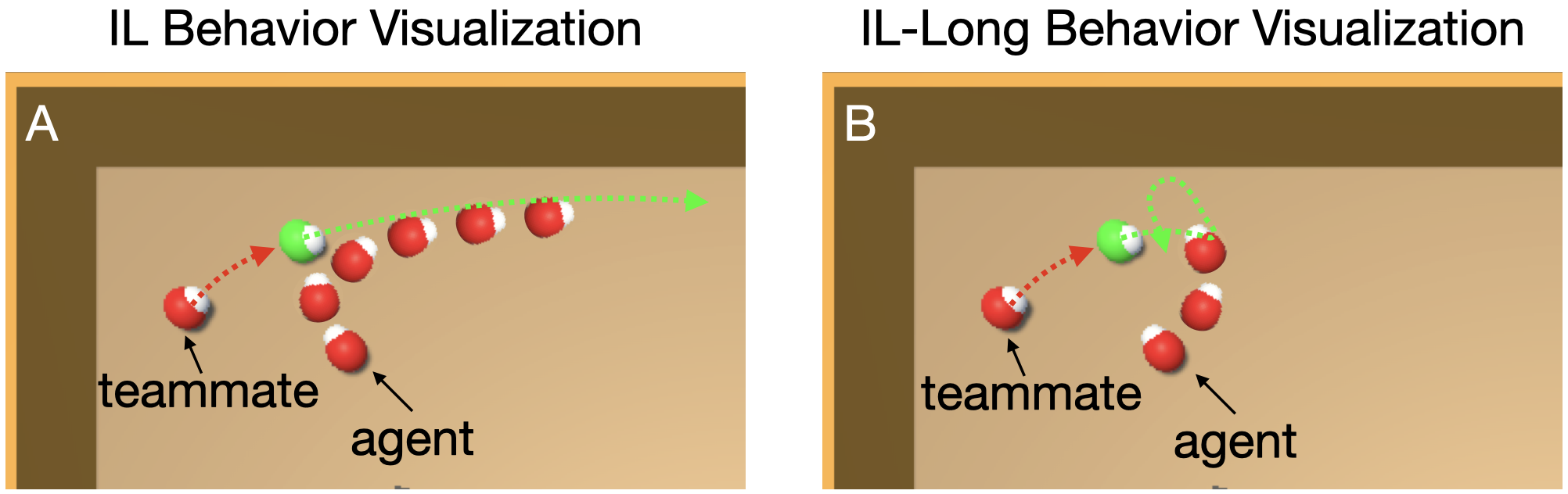}
    \caption{(A) IL behavior visualization. (B) IL-Long behavior visualization. }
    \label{fig:IL}
    \vspace*{-8mm}
\end{figure}

\mypara{Imitation Learning with Long-term Action Prediction (IL-Long)}
Standard IL uses the current agent observation $s_t$ as input to predict the next action $a_t$, which in our case corresponds to the agent's location in the next time step. However, even with a well-trained IL policy on $D_\text{heuristic}$, the seekers' performance is upper-bounded by the performance of the heuristic policy. We hypothesize that a strong collaborative policy must not only account for teamwork but also consider the opponent's behavior. Therefore, we enhance the IL policy by training it to predict long-term actions rather than only the next immediate step. Specifically, the policy predicts actions corresponding to the agent's $p_{t+5}$. We refer to this method as IL-Long. Intuitively, IL-Long allows seekers to implicitly anticipate the future movements of the hiders and adjust their actions accordingly. As a result, instead of simply reacting to a hider's detection, IL-Long enables the seeker to make more strategic decisions based on long-term predictions, as shown in Fig.~\ref{fig:IL}A-B.

\mypara{Fine-Tuning (FT) with Single-Human Guidance} During fine-tuning, to prevent overfitting on the limited $D_\text{human}$, we freeze the encoder and only update the MLP layers. Recall that $D_\text{human}$ contains both human interventions and heuristic-driven agent behaviors when no intervention is triggered. To mitigate catastrophic forgetting, we balance each training batch by including equal proportions of trajectories from each case.




\mypara{Policy Embedding} Fine-tuning on IL or IL-Long is insufficient for seekers to develop strong collaborative behaviors. A key limitation is that the policy encoder is primarily trained to predict heuristic seeker actions, which do not involve explicit collaboration. Inspired by ToM, we introduce a set of representations learning strategies to strengthen the policy encoder. The learned representations enable each seeker to consider their teammates' locations and behaviors, thereby fostering more effective collaboration. In addition, PE incorporates IL-Long to predict future positions, allowing seekers to anticipate opponent movements while collaborating with teammates. We refer to our method as Policy Embedding (PE). We follow similar training procedures as before: training on $D_\text{heuristic}$, freezing the encoder, and fine-tuning on $D_\text{human}$.

\begin{figure}[t!]
    \centering
    \includegraphics[width=0.45\textwidth]{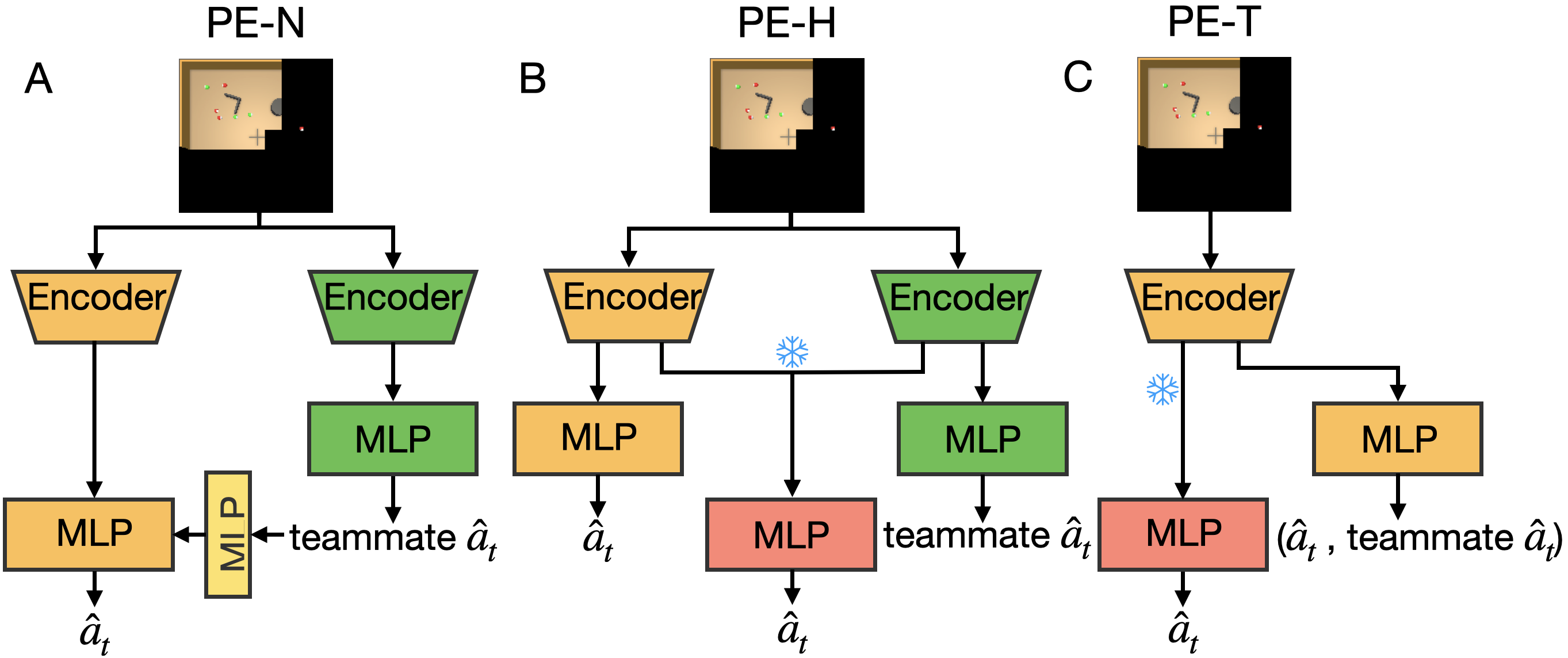}
    \vspace*{-2mm}
    \caption{Policy Embedding. (A) PE-N explicitly uses the predicted teammate actions as inputs for policy training. (B) PE-H combines the representation of teammate action prediction and self-action prediction for policy training. (C) PE-T learns effective representations by training one network to predict the actions of the whole team.}
    \label{fig:PE}
    \vspace*{-8mm}
\end{figure}

\mypara{Policy Embedding Naive (PE-N, Fig.~\ref{fig:PE}A)} involves training two separate networks. We first train a network to predict the actions of all seeker's teammates. We then train the second policy using IL-Long by incorporating the original visual observations along with the predicted teammate actions from the first network. The predicted actions are explicitly fed into the MLP of the IL-Long policy via a projection layer.

\mypara{Policy Embedding High-Level (PE-H, Fig.~\ref{fig:PE}B)} also trains two separate networks. However, instead of directly passing the predicted teammate actions into the MLP of the IL-Long policy, we freeze the encoders of both networks and concatenate the representations from each encoder. This combined representation is used to train a new MLP policy module with $D_\text{heuristic}$, which is then fine-tuned using $D_\text{human}$.

\mypara{Policy Embedding Team (PE-T, Fig.~\ref{fig:PE}C)} Both PE-N and PE-H aim to learn effective representations to capture teammate behaviors by training two separate networks. The key difference lies in how the learned representations are integrated into the main policy. However, training separate networks and then merging their representations can lead to representation alignment issues \cite{imani2022representation, jian2023policy}, where the learned representations may not seamlessly combine. In contrast, PE-T trains a single policy to predict the actions of the entire team. We find that PE-T provides the most effective representations. The learned encoder is then frozen, and an MLP module is trained with $D_\text{heuristic}$ and then $D_\text{human}$.

\mypara{Action Prediction Orders} When training policies to predict teammate actions, determining the order of action predictions is challenging because all seekers appear identical. In a dynamic task like multi-agent hide and seek, the relative positions of the seekers change significantly as the task progresses. To ensure stable training, it is crucial to establish a deterministic mapping between the input observations and the predicted actions. We propose a distance-based sorting mechanism, where the prediction order of all seekers is determined by their distance from the referenced seeker (i.e., the seeker associated with the input observation). This distance-based sorting ensures that the ordering of agents remains consistent with their spatial relationships for more effective reasoning.


\section{Experiments}

We conducted experiments in both simulation and the real-world. Throughout data collection, policy learning, and testing, we varied the environment layout as well as the number of seekers and hiders. Our goal is to evaluate whether our approach enables collaborative behaviors among seekers through single-human guidance and whether the learned policy generalizes across different settings. Additionally, we performed ablation studies and analyses to understand the effectiveness of our approach.


\begin{table}[h!]
\centering
\caption{Map and Agent Configurations}
\vspace*{-2mm}
\label{tab: platforms}
\resizebox{\columnwidth}{!}{
\footnotesize
\begin{tabular}{@{}c|c|c|c|c|c|c@{}}
\toprule
Domain  & Arena & Obstacle Type& \# Obstacles & Max & Speed & Visual
\\ &Size && & Time & & Range
\\ \midrule
Sim & 50$\times$50m& Cross, Rectangle, &5 & 120s & seeker(5m/s) & seeker(16m)
\\&&L-shape, Cylinder&  & & hider(8m/s) & hider(10m) \\
Real & 5$\times$5m& Cross, Rectangle & 2 & 60s & seeker(65rpm) & seeker(1.5m)
\\ & & & & & hider(130rpm) & hider(1.25m)
\\ \bottomrule

\end{tabular}
}



\vspace*{-6mm}
\end{table}

\subsection{Simulation Experiment}
We constructed our simulation using Unity engine and CREW \cite{zhang2024crew} to enable a single human to guide multiple seekers. Details regarding the map and agent configurations are in Table~\ref{tab: platforms}. For heuristic data collection, we gathered data from different settings ranging from 1 seeker vs. 1 hider, 2 seekers vs. 1 hider, and up to 4 seekers vs. 4 hiders across a total of 10 configurations. We ran 800 episodes per setting to collect training data. Each input frame has a dimension of $156 \times 156 \times 4$, including three RGB channels and one binary mask channel. Human expert data was collected for 10 episodes per setting, resulting in 40 minutes of data and 56k samples. During the intervention, the human expert selects which seeker to control and directs movement by clicking on future positions. We tested each learning policy in each setting with three random seeds and 150 episodes per seed.

\begin{table*}[t!]
\centering
\caption{Simulation Success Rate ($\%$)}
\vspace*{-2mm}
\label{tab: simulation}
\setlength{\tabcolsep}{2.5pt} 
\renewcommand{\arraystretch}{1} 
\footnotesize 
\begin{tabular}{l|c|c|cc|cc|ccc|ccc}
\toprule
Setting & \multicolumn{1}{c|}{2v1} & \multicolumn{1}{c|}{2v2} & \multicolumn{2}{c|}{3v2} & \multicolumn{2}{c}{3v3} & \multicolumn{3}{c}{4v3} & \multicolumn{3}{c}{4v4}
\\ \midrule

Heuristic & \multicolumn{1}{c|}{31.6$\pm$3.6} & \multicolumn{1}{c|}{23.3$\pm$3.6} & \multicolumn{2}{c|}{44.0$\pm$1.1} & \multicolumn{2}{c|}{36.4$\pm$4.1}& \multicolumn{3}{c|}{58.0$\pm$3.3} & \multicolumn{3}{c}{48.7$\pm$1.4} 
\\ \midrule

IL & \multicolumn{1}{c|}{17.1$\pm$3.1} & \multicolumn{1}{c|}{ 7.1$\pm$1.4} & \multicolumn{2}{c|}{ 16.4$\pm$1.3} & \multicolumn{2}{c|}{12.0$\pm$2.4} & \multicolumn{3}{c|}{23.8$\pm$3.5} & \multicolumn{3}{c}{19.1$\pm$3.0} \\

IL-Long & \multicolumn{1}{c|}{71.8$\pm$3.1} & \multicolumn{1}{c|}{55.3$\pm$2.0} & \multicolumn{2}{c|}{77.6$\pm$1.7} & \multicolumn{2}{c|}{66.4$\pm$0.8} & \multicolumn{3}{c|}{85.1$\pm$1.7} & \multicolumn{3}{c}{81.3$\pm$3.8 } \\

IL FT & \multicolumn{1}{c|}{18.9$\pm$2.3} & \multicolumn{1}{c|}{7.0$\pm$2.1} & \multicolumn{2}{c|}{28.9$\pm$1.7} & \multicolumn{2}{c|}{19.3$\pm$0.9} & \multicolumn{3}{c|}{38.7$\pm$2.8} & \multicolumn{3}{c}{14.4$\pm$1.7} \\

IL-Long FT & \multicolumn{1}{c|}{64.7$\pm$0.5} & \multicolumn{1}{c|}{46.2$\pm$3.5} & \multicolumn{2}{c|}{74.7$\pm$0.5} & \multicolumn{2}{c|}{66.0$\pm$5.0} & \multicolumn{3}{c|}{88.0$\pm$2.0} & \multicolumn{3}{c}{80.7$\pm$1.4} \\
PE-N & \multicolumn{1}{c|}{59.6$\pm$3.6} & \multicolumn{1}{c|}{46.4$\pm$1.6} & \multicolumn{2}{c|}{75.6$\pm$0.8} & \multicolumn{2}{c|}{51.3$\pm$2.4} & \multicolumn{3}{c|}{88.4$\pm$0.6} & \multicolumn{3}{c}{73.6$\pm$2.1}\\
PE-H & \multicolumn{1}{c|}{71.8$\pm$3.2} & \multicolumn{1}{c|}{51.6$\pm$4.0} & \multicolumn{2}{c|}{70.2$\pm$0.6} & \multicolumn{2}{c|}{58.2$\pm$2.1} & \multicolumn{3}{c|}{84.9$\pm$3.3} & \multicolumn{3}{c}{81.6$\pm$3.6}\\
PE-T & \multicolumn{1}{c|}{78.7$\pm$1.9} & \multicolumn{1}{c|}{67.3$\pm$2.9} & \multicolumn{2}{c|}{90.9$\pm$1.4} & \multicolumn{2}{c|}{\textbf{86.0$\pm$4.3}} & \multicolumn{3}{c|}{94.7$\pm$2.7} & \multicolumn{3}{c}{\textbf{94.2$\pm$1.4}}\\
\midrule
Combination & 1+1  & 1+1  & 2+1 & 1+2 & 2+1 & 1+2 & 3+1 & 2+2 & 1+3 & 3+1 & 2+2 & 1+3
\\ \midrule

IL-Long+IL-Long FT & 71.8$\pm$0.8 & 53.1$\pm$5.1 & 83.1$\pm$2.3 & 86.0$\pm$1.4 & 74.9$\pm$0.8 & 74.7$\pm$2.0 & 92.4$\pm$1.3 & 88.7$\pm$0.0 & 92.4$\pm$1.1 & 87.1$\pm$1.4 & 89.6$\pm$3.0 & 87.1$\pm$2.3 \\
IL-Long+PE-N & 74.4$\pm$1.3 & 50.0$\pm$4.3 & 83.3$\pm$2.0 & 86.2$\pm$0.6 & 76.2$\pm$2.5 & 77.3$\pm$2.4 & 89.8$\pm$1.4 & 92.9$\pm$2.1 & 90.9$\pm$0.8 & 89.8$\pm$3.0 & 90.4$\pm$0.3 & 88.0$\pm$0.9 \\
IL-Long+PE-H & 84.2$\pm$0.8 & 71.3$\pm$0.5 & 87.8$\pm$2.5 & 85.6$\pm$4.2 & 77.3$\pm$5.0 & 76.0$\pm$1.9 & \textbf{94.2$\pm$2.5} & 93.3$\pm$0.5 & 93.1$\pm$0.8 & 90.0$\pm$2.5 & 92.4$\pm$1.7 & 89.8$\pm$3.3 \\
IL-Long+PE-T & \textbf{89.3$\pm$2.0} & \textbf{72.2$\pm$3.3} & \textbf{91.3$\pm$2.4} & \textbf{94.9$\pm$1.4} & \textbf{83.1$\pm$1.7} & \textbf{86.2$\pm$0.6} & 90.9$\pm$1.1 & \textbf{96.4$\pm$1.7} & \textbf{96.2$\pm$0.6} & \textbf{92.0$\pm$2.0} & \textbf{93.6$\pm$1.4} & \textbf{93.3$\pm$0.9} \\

\bottomrule
\end{tabular}
\vspace*{-5mm}
\end{table*}

We present the results in Table~\ref{tab: simulation}. Seekers with the heuristic policy perform poorly in all settings, indicating that collaboration is essential for success in this task. IL performs worse than the heuristic due to the inherent limitation of the training data. In contrast, IL-Long significantly outperforms both IL and the heuristic, which implies that predicting several steps ahead helps the policy implicitly model the hiders' behaviors. Fine-tuning IL and IL-Long does not lead to improvements, likely because simply imitating human interventions without capturing the underlying reasoning mechanisms from a small set of data is challenging. Interestingly, explicitly predicting all teammate actions, as in PE-N, does not improve performance, suggesting that the right level representation is not captured. However, PE-H, which incorporates the latent representation of a teammate action prediction model into policy learning, outperforms PE-N in most settings. Finally, PE-T, which learns representations by predicting the actions of all team members and uses these for fine-tuning, significantly outperforms all other methods across all settings. Jointly learning to predict both the agent's own actions and those of its teammates provides the most effective representations for grounding human guidance.

Additionally, we experimented with different combinations of policies among seekers. As discussed before, IL-Long produces the most effective chasing behavior but no collaborations while PE-T shows the best collaborative behavior. Therefore, teams of seekers combining IL and PE-T generally achieve the highest performance.

\subsection{Real-World Experiment}
Our real-world experiments aim to test if our approach scales well to learning under noisy conditions such as actuation noises and imperfect perception. All agents were modified from DJI RoboMaster S1 vehicles. Partial visual observations were configured similarly to our simulation using a top-down camera and a motion capture system. The map and robot configurations are reported in Table~\ref{tab: platforms}. We collected 50 episodes each for 4 settings (2 seekers v 1 hiders, 2 seekers v 2 hiders, 3 seekers v 2 hiders, and 3 seekers v 3 hiders). For human guidance, we collected 20 episodes per setting, totaling 45 minutes of data and 36k samples. We tested each setting with 20 episodes in Table~\ref{tab: real}. The results mirror those in our simulation, showing the scalability of our approach in real-world multi-robot systems.

\begin{table}[t!]
\centering
\caption{Real-World Success Rate (success/total trails)}
\label{tab: real}
\setlength{\tabcolsep}{5pt} 
\renewcommand{\arraystretch}{1} 
\footnotesize 
\resizebox{0.425\textwidth}{!}{%
\begin{tabular}{l|c|c|cc|cc}
\toprule
Setting & \multicolumn{1}{c|}{2v1} & \multicolumn{1}{c|}{2v2} &\multicolumn{2}{c|}{3v2} & \multicolumn{2}{c}{3v3} 
\\ \midrule
Heuristic &\multicolumn{1}{c|}{11/20} & \multicolumn{1}{c|}{10/20}& \multicolumn{2}{c|}{13/20} & \multicolumn{2}{c}{11/20} \\
\midrule 
IL-Long &\multicolumn{1}{c|}{10/20} & \multicolumn{1}{c|}{7/20} & \multicolumn{2}{c|}{15/20} & \multicolumn{2}{c}{9/20}\\
\midrule
Combination & 1+1 & 1+1 & 2+1 & 2+1&2+1&1+2\\
\midrule
IL-Long+PE-T& \textbf{12/20}& \textbf{11/20} & \textbf{16/20} &  \textbf{17/20}&\textbf{16/20}& \textbf{16/20}\\
\bottomrule
\end{tabular}
}
\vspace*{-3mm}
\end{table}

\subsection{Analysis}

\begin{figure}[t!]
    \centering
    \includegraphics[width=0.45\textwidth]{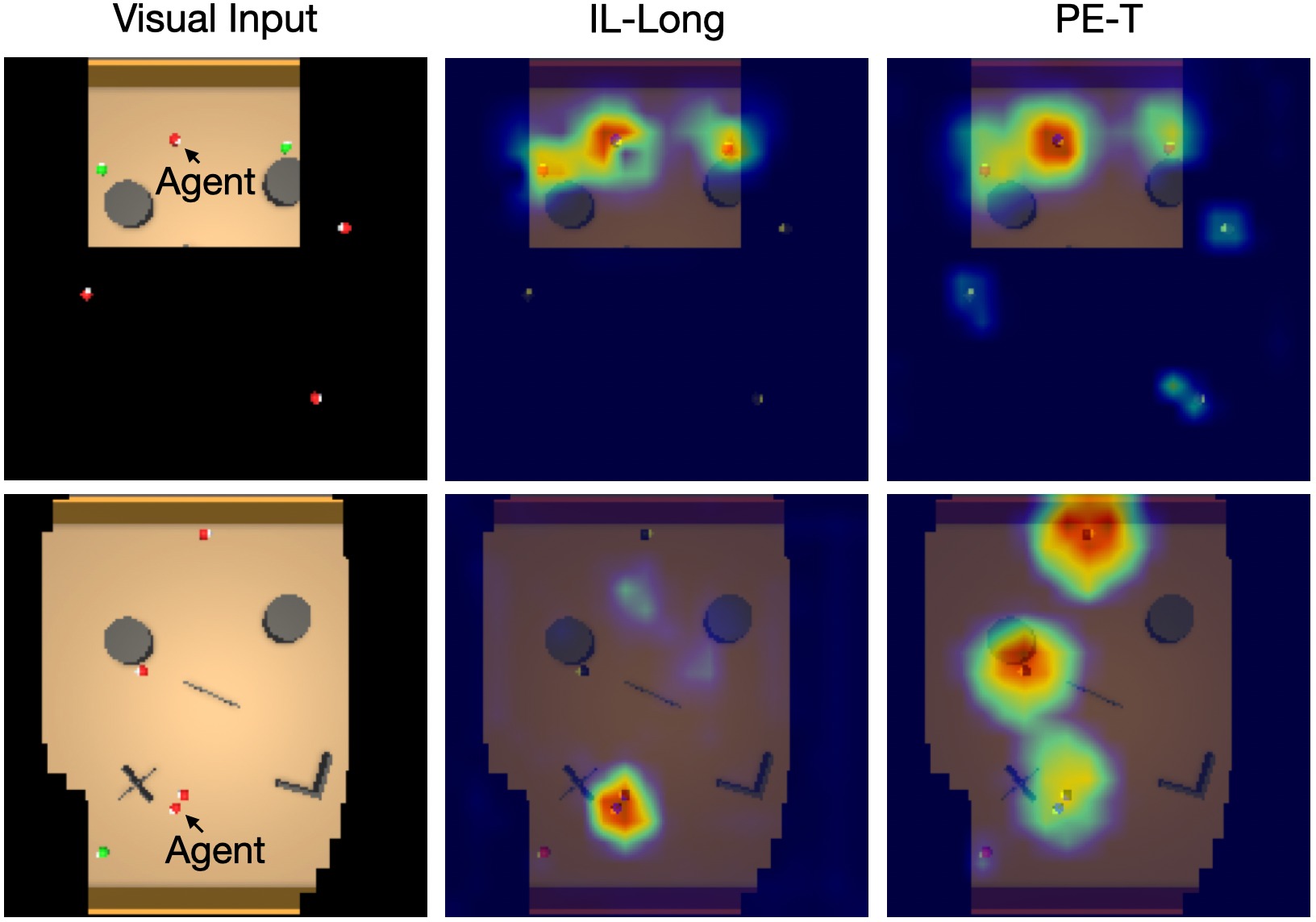}
    \caption{Policy attention map with Grad-CAM.}
    \label{fig:Grad}
    \vspace*{-6mm}
\end{figure}

We hypothesize that our policy embedding methods enable stronger collaborations because these policies learn effective representations to capture both teammate and opponent behaviors. By visualizing the attention map of the policies using Grad-CAM \cite{selvaraju2017grad} in Fig.~\ref{fig:Grad}, while IL-Long primarily focuses on the seeker itself and the visible hiders, our PE-T attends to both the visible hiders and their teammates.

We also quantitatively assessed how much a policy relies on the information about its teammates. For instance, in 2 seekers vs. 1 hider and 2 seekers vs. 2 hiders settings, where one seeker is controlled by the heuristic policy, the other seeker must collaborate with it to succeed. In these cases, when we removed the teammate position encoding from the visual inputs, the performance of the seeker team controlled by heuristic and IL-Long remained unaffected. However, the performance of the seeker team controlled by heuristic and PE-T drops significantly, from $78.67\%$ to $66\%$ and from $56.44\%$ to $48.22\%$, respectively.

    

Additionally, we examined how the duration of human intervention affects the fine-tuned policy. We tested on the two settings mentioned above. Among two seekers, one was controlled by heuristic policy, and the other one was controlled by PE-T. Our results indicate that fine-tuning with only 10 minutes of human guidance already improved the success rate of IL-Long by $22.1\%$ and $29.2\%$. With 20 minutes of human guidance, the success rate increased by $28.2\%$ and $46.7\%$. Finally, with 40 minutes of human guidance, the success rate improved by $42.6\%$ and $59.8\%$.

    

\section{Conclusion, Limitation, and Future Work}

We introduce a novel framework to enable multi-agent collaboration from single-human guidance. Our method alleviates demonstrations from a team of experts in MAIL and tackles the uncertainty about when and if collaborative behavior will emerge in traditional MARL. Inspired by Theory of Mind, we also develop a policy embedding method to learn effective representations of teammate and opponent behaviors for grounding limited human guidance data. In challenging multi-agent hide-and-seek tasks, our method outperformed baselines by up to $58\%$ with strong collaborative behaviors. Additionally, our approach successfully transferred to real-world multi-robot systems with robust generalization across domains.

As for the limitations, currently, human guidance is limited to clicking to specify an agent's future actions, but other modalities \cite{zhang2024guide}, such as languages, could be investigated to enhance interaction. Additionally, exploring how a small group of humans can effectively guide a large swarm of agents presents an exciting opportunity to scale up our system.

\renewcommand*{\bibfont}{\small}
\bibliographystyle{IEEEtran}
\bibliography{reference}

\begin{thebibliography}{10}
\providecommand{\url}[1]{#1}
\csname url@samestyle\endcsname
\providecommand{\newblock}{\relax}
\providecommand{\bibinfo}[2]{#2}
\providecommand{\BIBentrySTDinterwordspacing}{\spaceskip=0pt\relax}
\providecommand{\BIBentryALTinterwordstretchfactor}{4}
\providecommand{\BIBentryALTinterwordspacing}{\spaceskip=\fontdimen2\font plus
\BIBentryALTinterwordstretchfactor\fontdimen3\font minus \fontdimen4\font\relax}
\providecommand{\BIBforeignlanguage}[2]{{%
\expandafter\ifx\csname l@#1\endcsname\relax
\typeout{** WARNING: IEEEtran.bst: No hyphenation pattern has been}%
\typeout{** loaded for the language `#1'. Using the pattern for}%
\typeout{** the default language instead.}%
\else
\language=\csname l@#1\endcsname
\fi
#2}}
\providecommand{\BIBdecl}{\relax}
\BIBdecl

\bibitem{huamncolab}
A.~P. Melis, ``The evolutionary roots of human collaboration: coordination and sharing of resources,'' \emph{Annals of the New York Academy of Sciences}, vol. 1299, no.~1, pp. 68--76, 2013.

\bibitem{maddpg}
R.~Lowe, Y.~I. Wu, A.~Tamar, J.~Harb, O.~Pieter~Abbeel, and I.~Mordatch, ``Multi-agent actor-critic for mixed cooperative-competitive environments,'' \emph{Advances in neural information processing systems}, vol.~30, 2017.

\bibitem{maac}
S.~Iqbal and F.~Sha, ``Actor-attention-critic for multi-agent reinforcement learning,'' in \emph{International conference on machine learning}.\hskip 1em plus 0.5em minus 0.4em\relax PMLR, 2019, pp. 2961--2970.

\bibitem{mappo}
C.~Yu, A.~Velu, E.~Vinitsky, J.~Gao, Y.~Wang, A.~Bayen, and Y.~Wu, ``The surprising effectiveness of ppo in cooperative multi-agent games,'' \emph{Advances in Neural Information Processing Systems}, vol.~35, pp. 24\,611--24\,624, 2022.

\bibitem{qmix}
T.~Rashid, M.~Samvelyan, C.~S. De~Witt, G.~Farquhar, J.~Foerster, and S.~Whiteson, ``Monotonic value function factorisation for deep multi-agent reinforcement learning,'' \emph{Journal of Machine Learning Research}, vol.~21, no. 178, pp. 1--51, 2020.

\bibitem{vdn}
P.~Sunehag, G.~Lever, A.~Gruslys, W.~M. Czarnecki, V.~Zambaldi, M.~Jaderberg, M.~Lanctot, N.~Sonnerat, J.~Z. Leibo, K.~Tuyls \emph{et~al.}, ``Value-decomposition networks for cooperative multi-agent learning,'' \emph{arXiv preprint arXiv:1706.05296}, 2017.

\bibitem{tampuu2017multiagent}
A.~Tampuu, T.~Matiisen, D.~Kodelja, I.~Kuzovkin, K.~Korjus, J.~Aru, J.~Aru, and R.~Vicente, ``Multiagent cooperation and competition with deep reinforcement learning,'' \emph{PloS one}, vol.~12, no.~4, p. e0172395, 2017.

\bibitem{socialdilemma}
J.~Z. Leibo, V.~Zambaldi, M.~Lanctot, J.~Marecki, and T.~Graepel, ``Multi-agent reinforcement learning in sequential social dilemmas,'' \emph{arXiv preprint arXiv:1702.03037}, 2017.

\bibitem{emergethrucompete}
\BIBentryALTinterwordspacing
S.~Liu, G.~Lever, N.~Heess, J.~Merel, S.~Tunyasuvunakool, and T.~Graepel, ``Emergent coordination through competition,'' in \emph{International Conference on Learning Representations}, 2019. [Online]. Available: \url{https://openreview.net/forum?id=BkG8sjR5Km}
\BIBentrySTDinterwordspacing

\bibitem{capture}
M.~Jaderberg, W.~Czarnecki, I.~Dunning, T.~Graepel, and L.~Marris, ``Capture the flag: the emergence of complex cooperative agents,'' \emph{DeepMind blog. May}, vol.~30, 2019.

\bibitem{bakeremergent}
\BIBentryALTinterwordspacing
B.~Baker, I.~Kanitscheider, T.~Markov, Y.~Wu, G.~Powell, B.~McGrew, and I.~Mordatch, ``Emergent tool use from multi-agent autocurricula,'' in \emph{International Conference on Learning Representations}, 2020. [Online]. Available: \url{https://openreview.net/forum?id=SkxpxJBKwS}
\BIBentrySTDinterwordspacing

\bibitem{le2017coordinated}
H.~M. Le, Y.~Yue, P.~Carr, and P.~Lucey, ``Coordinated multi-agent imitation learning,'' in \emph{International Conference on Machine Learning}.\hskip 1em plus 0.5em minus 0.4em\relax PMLR, 2017, pp. 1995--2003.

\bibitem{copulas}
H.~Wang, L.~Yu, Z.~Cao, and S.~Ermon, ``Multi-agent imitation learning with copulas,'' in \emph{Machine Learning and Knowledge Discovery in Databases. Research Track: European Conference, ECML PKDD 2021, Bilbao, Spain, September 13--17, 2021, Proceedings, Part I 21}.\hskip 1em plus 0.5em minus 0.4em\relax Springer, 2021, pp. 139--156.

\bibitem{driving}
R.~P. Bhattacharyya, D.~J. Phillips, B.~Wulfe, J.~Morton, A.~Kuefler, and M.~J. Kochenderfer, ``Multi-agent imitation learning for driving simulation,'' in \emph{2018 IEEE/RSJ International Conference on Intelligent Robots and Systems (IROS)}.\hskip 1em plus 0.5em minus 0.4em\relax IEEE, 2018, pp. 1534--1539.

\bibitem{multigail}
J.~Song, H.~Ren, D.~Sadigh, and S.~Ermon, ``Multi-agent generative adversarial imitation learning,'' \emph{Advances in neural information processing systems}, vol.~31, 2018.

\bibitem{scalable}
W.~Jeon, P.~Barde, D.~Nowrouzezahrai, and J.~Pineau, ``Scalable and sample-efficient multi-agent imitation learning,'' in \emph{Proceedings of the Workshop on Artificial Intelligence Safety, co-located with 34th AAAI Conference on Artificial Intelligence, SafeAI@ AAAI}, 2020.

\bibitem{qin2025perception}
Y.~Qin, R.~T. Lee, and P.~Sajda, ``Perception of an ai teammate in an embodied control task affects team performance, reflected in human teammates' behaviors and physiological responses,'' \emph{arXiv preprint arXiv:2501.15332}, 2025.

\bibitem{qin2025physiologically}
Y.~Qin, R.~T. Lee, W.~Zhang, X.~Sun, and P.~Sajda, ``Physiologically-informed predictability of a teammate's future actions forecasts team performance,'' \emph{arXiv preprint arXiv:2501.15328}, 2025.

\bibitem{wellman2004scaling}
H.~M. Wellman and D.~Liu, ``Scaling of theory-of-mind tasks,'' \emph{Child development}, vol.~75, no.~2, pp. 523--541, 2004.

\bibitem{apperly2009humans}
I.~A. Apperly and S.~A. Butterfill, ``Do humans have two systems to track beliefs and belief-like states?'' \emph{Psychological review}, vol. 116, no.~4, p. 953, 2009.

\bibitem{dqn}
V.~Mnih, K.~Kavukcuoglu, D.~Silver, A.~A. Rusu, J.~Veness, M.~G. Bellemare, A.~Graves, M.~Riedmiller, A.~K. Fidjeland, G.~Ostrovski \emph{et~al.}, ``Human-level control through deep reinforcement learning,'' \emph{nature}, vol. 518, no. 7540, pp. 529--533, 2015.

\bibitem{ppo}
J.~Schulman, F.~Wolski, P.~Dhariwal, A.~Radford, and O.~Klimov, ``Proximal policy optimization algorithms,'' \emph{arXiv preprint arXiv:1707.06347}, 2017.

\bibitem{ddpg}
T.~Lillicrap, ``Continuous control with deep reinforcement learning,'' \emph{arXiv preprint arXiv:1509.02971}, 2015.

\bibitem{sac}
T.~Haarnoja, A.~Zhou, P.~Abbeel, and S.~Levine, ``Soft actor-critic: Off-policy maximum entropy deep reinforcement learning with a stochastic actor,'' in \emph{International conference on machine learning}.\hskip 1em plus 0.5em minus 0.4em\relax PMLR, 2018, pp. 1861--1870.

\bibitem{reviewcoop}
Y.~Du, J.~Z. Leibo, U.~Islam, R.~Willis, and P.~Sunehag, ``A review of cooperation in multi-agent learning,'' \emph{CoRR}, 2023.

\bibitem{yang2020bayesian}
F.~Yang, A.~Vereshchaka, C.~Chen, and W.~Dong, ``Bayesian multi-type mean field multi-agent imitation learning,'' \emph{Advances in Neural Information Processing Systems}, vol.~33, pp. 2469--2478, 2020.

\bibitem{agentsmodelagents}
S.~V. Albrecht and P.~Stone, ``Autonomous agents modelling other agents: A comprehensive survey and open problems,'' \emph{Artificial Intelligence}, vol. 258, pp. 66--95, 2018.

\bibitem{machinetom}
N.~Rabinowitz, F.~Perbet, F.~Song, C.~Zhang, S.~A. Eslami, and M.~Botvinick, ``Machine theory of mind,'' in \emph{International conference on machine learning}.\hskip 1em plus 0.5em minus 0.4em\relax PMLR, 2018, pp. 4218--4227.

\bibitem{cogmachinetom}
T.~N. Nguyen and C.~Gonzalez, ``Cognitive machine theory of mind.'' in \emph{CogSci}, 2020.

\bibitem{chen2021visualtob}
B.~Chen, C.~Vondrick, and H.~Lipson, ``Visual behavior modelling for robotic theory of mind,'' \emph{Scientific Reports}, vol.~11, no.~1, p. 424, 2021.

\bibitem{chen2021visual}
B.~Chen, Y.~Hu, R.~Kwiatkowski, S.~Song, and H.~Lipson, ``Visual perspective taking for opponent behavior modeling,'' in \emph{2021 IEEE International Conference on Robotics and Automation (ICRA)}.\hskip 1em plus 0.5em minus 0.4em\relax IEEE, 2021, pp. 13\,678--13\,685.

\bibitem{symmetricmachinetom}
M.~Sclar, G.~Neubig, and Y.~Bisk, ``Symmetric machine theory of mind,'' in \emph{International Conference on Machine Learning}.\hskip 1em plus 0.5em minus 0.4em\relax PMLR, 2022, pp. 19\,450--19\,466.

\bibitem{hideandseek}
B.~Chen, S.~Song, H.~Lipson, and C.~Vondrick, ``Visual hide and seek,'' in \emph{Artificial Life Conference Proceedings 32}.\hskip 1em plus 0.5em minus 0.4em\relax MIT Press One Rogers Street, Cambridge, MA 02142-1209, USA journals-info~…, 2020, pp. 645--655.

\bibitem{smith1986representation}
R.~C. Smith and P.~Cheeseman, ``On the representation and estimation of spatial uncertainty,'' \emph{The international journal of Robotics Research}, vol.~5, no.~4, pp. 56--68, 1986.

\bibitem{He_2016_CVPR}
K.~He, X.~Zhang, S.~Ren, and J.~Sun, ``Deep residual learning for image recognition,'' in \emph{Proceedings of the IEEE Conference on Computer Vision and Pattern Recognition (CVPR)}, June 2016.

\bibitem{imani2022representation}
\BIBentryALTinterwordspacing
E.~Imani, W.~Hu, and M.~White, ``Representation alignment in neural networks,'' \emph{Transactions on Machine Learning Research}, 2022. [Online]. Available: \url{https://openreview.net/forum?id=fLIWMnZ9ij}
\BIBentrySTDinterwordspacing

\bibitem{jian2023policy}
P.~Jian, E.~Lee, Z.~Bell, M.~M. Zavlanos, and B.~Chen, ``Policy stitching: Learning transferable robot policies,'' in \emph{Conference on Robot Learning}.\hskip 1em plus 0.5em minus 0.4em\relax PMLR, 2023, pp. 3789--3808.

\bibitem{zhang2024crew}
\BIBentryALTinterwordspacing
L.~Zhang, Z.~Ji, and B.~Chen, ``{CREW}: Facilitating human-{AI} teaming research,'' \emph{Transactions on Machine Learning Research}, 2024. [Online]. Available: \url{https://openreview.net/forum?id=ZRXwHRXm8i}
\BIBentrySTDinterwordspacing

\bibitem{selvaraju2017grad}
R.~R. Selvaraju, M.~Cogswell, A.~Das, R.~Vedantam, D.~Parikh, and D.~Batra, ``Grad-cam: Visual explanations from deep networks via gradient-based localization,'' in \emph{Proceedings of the IEEE international conference on computer vision}, 2017, pp. 618--626.

\bibitem{zhang2024guide}
L.~Zhang, Z.~Ji, N.~R. Waytowich, and B.~Chen, ``{GUIDE}: Real-time human-shaped agents,'' \emph{Advances in Neural Information Processing Systems}, vol.~38, 2024.

\end{thebibliography}

\end{document}